\documentclass[runningheads]{llncs}

\usepackage[mobile]{eccv}

\usepackage{eccvabbrv}

\usepackage{graphicx}
\usepackage{booktabs}

\usepackage{amsthm,amsmath,amssymb}
\usepackage{mathrsfs}
\usepackage{multirow}
\usepackage{makecell}
\usepackage{diagbox}
\usepackage{colortbl}
\usepackage[figuresright]{rotating}

\usepackage[accsupp]{axessibility}  

\usepackage{hyperref}

\usepackage{orcidlink}

\begin{document}

\title{MoSA: Mixture of Sparse Adapters for \protect\\ Visual Efficient Tuning} 


\author{\small Qizhe Zhang\inst{1} \and
Bocheng Zou\inst{2} \and
Ruichuan An\inst{3} \and
Jiaming Liu\inst{1} \and
Shanghang Zhang\inst{1}}

\authorrunning{MoSA: Mixture of Sparse Adapters for Visual Efficient Tuning}

\institute{National Key Laboratory for Multimedia Information Processing, \\
School of Computer Science, Peking University \\
\email{\{theia, jiamingliu, shanghang\}@pku.edu.cn} \and
University of Wisconsin-Madison \\
\email{\{bochengz\}@cs.wisc.edu}\and
School of Software Engineering, Xi’an Jiaotong University \\
\email{\{arctanx\}@stu.xjtu.edu.cn}}

\maketitle

\begin{abstract}
With the rapid growth in the scale of pre-trained foundation models, parameter-efficient fine-tuning techniques have gained significant attention, among which Adapter Tuning is the most widely used. Despite achieving efficiency, it still underperforms full fine-tuning, and the performance improves at the cost of an increase in parameters. Recent efforts have either focused on training multiple adapter experts to increase model capacity or on pruning adapters to achieve parameter efficiency. However, both approaches introduce more parameters compared to the original adapter, hence are not computationally efficient. Motivated by this, we propose \textbf{M}ixture \textbf{o}f \textbf{S}parse \textbf{A}dapters, or \textbf{MoSA}, as a novel Adapter Tuning method to fully unleash the potential of each parameter in the adapter. We first split the standard adapter into multiple non-overlapping modules, then stochastically activate them for sparse training, and finally merge them to form a complete adapter after tuning. In this way, MoSA can achieve significantly better performance than standard adapters without any additional computational or storage overhead. Furthermore, we propose a hierarchical sparse strategy to better leverage limited training data. Extensive experiments on a series of 27 visual tasks demonstrate that MoSA consistently outperforms other Adapter Tuning methods as well as other baselines by a large margin. Furthermore, MoSA brings consistent improvements across various model scales, architectures, and different PEFT methods. Code will be released.
  \keywords{Adapter Tuning \and Mixture-of-Experts \and Sparse Training}
\end{abstract}

\section{Introduction}
\label{sec:intro}

\begin{figure}[t]
    \centering
    \begin{subfigure}{0.48\linewidth}
        \centering
        \includegraphics[width=\linewidth]{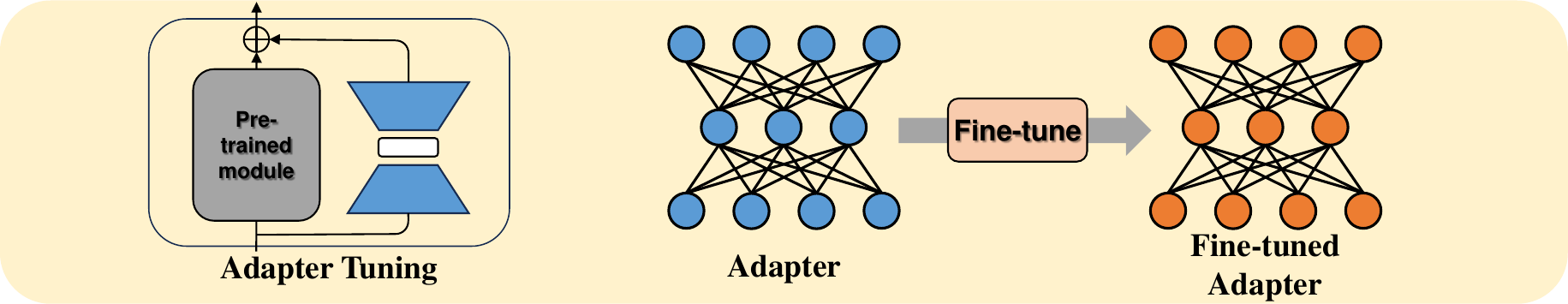}
        \caption{Standard Adapter}
        \label{fig:schema-adapter}
    \end{subfigure}
    \hfill
    \begin{subfigure}{0.48\linewidth}
        \centering
        \includegraphics[height=0\linewidth]{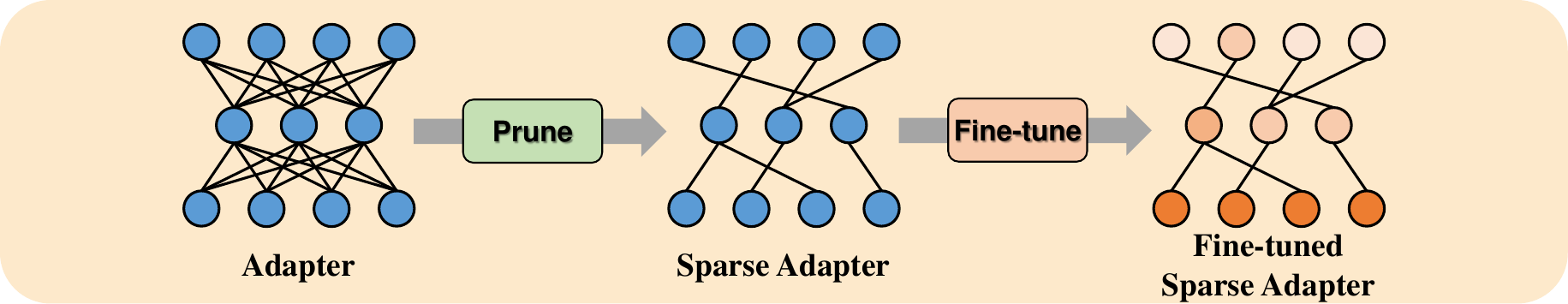}
    \end{subfigure}
    \vspace{-16mm}
    \newline
    \begin{subfigure}{0.48\linewidth}
        \centering
        \includegraphics[width=\linewidth]{fig/SparseAdapter.pdf}
        \caption{Sparse Adapter}
        \label{fig:schema-sparseadapter}
    \end{subfigure}
    \hfill
    \begin{subfigure}{0.48\linewidth}
        \centering
        \includegraphics[width=\linewidth]{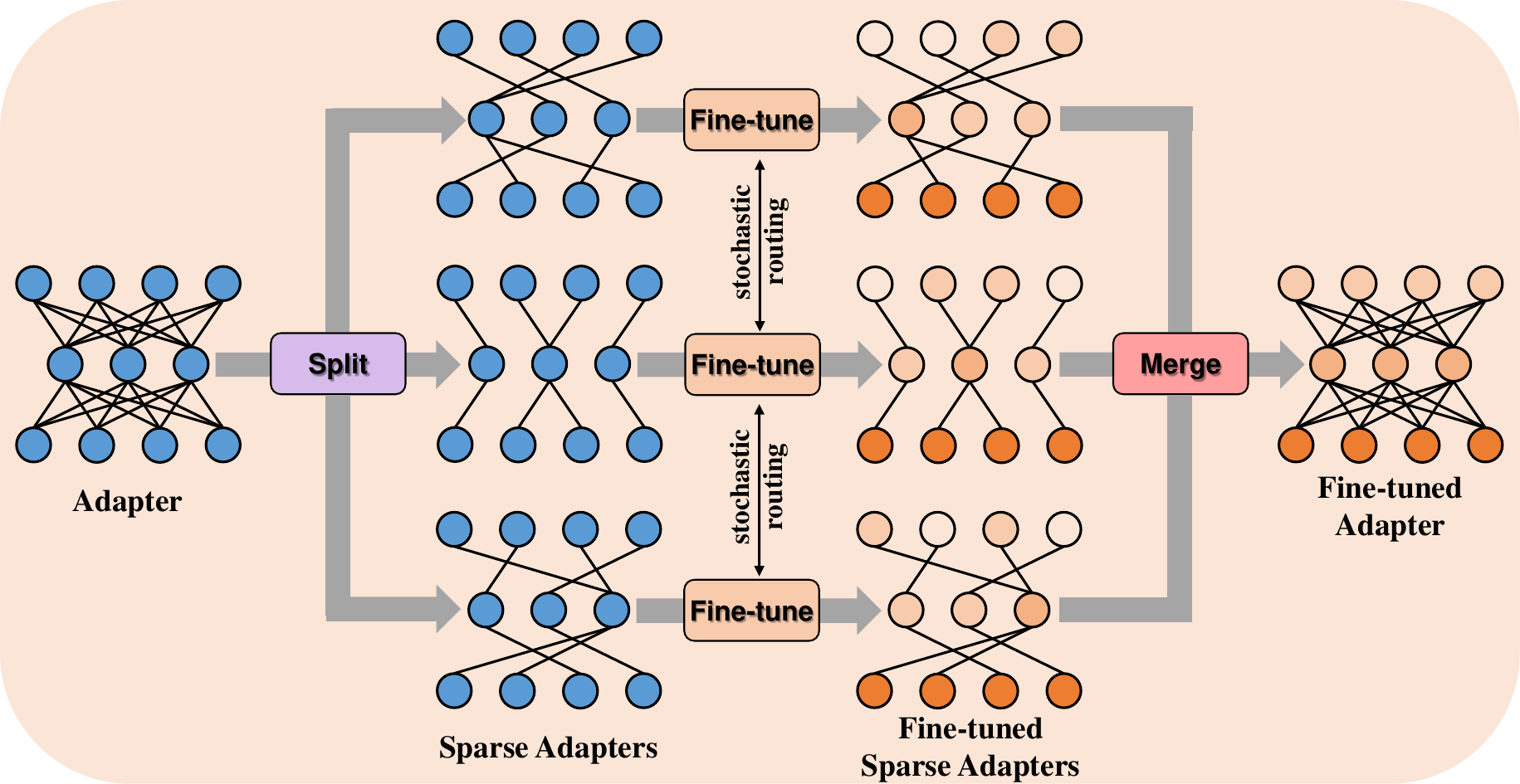}
        \caption{Mixture of Sparse Adapters}
        \label{fig:schema-mosa}
    \end{subfigure}
    \caption{Different Adapter Tuning diagrams: (a) The standard adapter simply inserts a bottleneck module into each Transformer layer. (b) Sparse Adapter. It prunes the standard adapter before tuning, updating only a small subset of the retained parameters. (c) Our proposed Mixture of Sparse Adapters (MoSA). We first split the standard adapter into multiple non-overlapping modules, then stochastically activate them for sparse training, and finally merge them to form a complete dense adapter.}
    \label{fig:schema}
\end{figure}

The \textit{pretrain-then-finetune} paradigm has achieved remarkable success in deep learning. Within the field of computer vision, models pre-trained on large-scale datasets (\eg ImageNet-21k~\cite{russakovsky2015imagenet}, JFT-300M~\cite{sun2017jft}, SA-1B~\cite{kirillov2023sam}) have demonstrated significant performance improvements across various downstream tasks~\cite{chen2020simclr, he2020moco, he2022mae}. After pre-training, models require fine-tuning on specific data to transfer learned knowledge to the target domain. The most direct method is \textit{full fine-tuning}, involving the update of all parameters in the pre-trained model during tuning. However, as the scale of pre-trained models continues to grow (\eg ViT-G/14~\cite{zhai2022vitg} 1.8B, LVM~\cite{bai2023lvm} 3B), storing a complete copy of all parameters for each task becomes impractical, giving rise to a more efficient method of tuning, known as \textit{parameter-efficient fine-tuning} (PEFT). For each downstream task, PEFT updates only a small portion of the the parameters in the pre-trained model, achieving efficiency in parameter storage and data utilization.

Recently, various PEFT methods have emerged~\cite{zaken2021bitfit, li2021prefix, lester2021prompt, liu2022ptuning, jia2022vpt, karimi2021compacter, liu2022ia3, hu2021lora, lian2022ssf, zhang2023gps, he2022kadaptation}, with one of the most widely used being \textit{Adapter Tuning}~\cite{houlsby2019adapter, pfeiffer2020adapterfusion, pfeiffer2020adapterhub, chen2022adaptformer, chen2023samadapter, pan2022stadapter, park2023dualpath}. This kind of method introduces lightweight bottleneck modules to the pre-trained model while freezing the backbone, as shown in \Cref{fig:schema}(a), facilitating efficient knowledge transfer to downstream tasks. Despite the respectable performance of these methods, they still fall behind full fine-tuning in many scenarios~\cite{han2024elephant} (\eg training data is relatively abundant, the distribution gap between the downstream and pre-training data is significant). Simply increasing the bottleneck dimension can raise performance, but this contradicts the original design philosophy of Adapter Tuning. Recent work has enhanced adapter performance by incorporating a Mixture-of-Experts (MoE) mechanism~\cite{zadouri2023moepeft, gou2023mocle, dou2023loramoe, gao2024mola}, but having multiple parallel adapter modules and routers introduces additional parameters and computation. Sparsely or stochastically activated MoE is more efficient in implementation~\cite{gao2024mola, chen2024llavamole, wang2022adamix}, but this reduces the amount of data seen by individual experts, an issue we call \textit{data dilution}, leading to even worse performance, especially when training data is limited. 
Meanwhile, another way for improving adapters involves pruning before tuning~\cite{he2022sparseadapter} (\eg a $4\times$ larger adapter with a 75\% sparse ratio), as in \Cref{fig:schema}(b), which has shown performance gains but does not offer computational efficiency, and high levels of sparsity can further lead to training instability. The delicate balance between parameter efficiency and performance remains a key challenge in Adapter Tuning.

Therefore, a natural question arises: \textbf{is it possible to achieve both efficiency and performance simultaneously}? In this paper, we propose \textbf{M}ixture \textbf{o}f \textbf{S}parse \textbf{A}dapters (\textbf{MoSA}) as an affirmative answer to this question, enhancing adapter performance without introducing additional computational and storage overhead, which is shown in \Cref{fig:schema}(c). We start by splitting the standard adapter into several non-overlapping modules, each can be considered as a sparse expert. 
During training, to avoid extra computational costs, we use a non-routing stochastic activation mechanism. Each activated module uses a mask to filter gradients, achieving sparsity in parameter updates. We merge all sparse adapters into a complete dense adapter, achieving efficiency in both storage and computation during inference. Through this design, the potential of the original adapter is fully unleashed, maximizing the elimination of parameter redundancy. 

Rather than simply stacking two independent approaches, our design organically integrates the stochastically activated MoE with sparse training, where the two components mutually reinforce each other. On one hand, sparse update of parameters more efficiently utilizes training data, alleviating the data dilution caused by sparse activation of multiple experts. On the other hand, the training process with mixed experts ensures expressive capability and stability in downstream tasks, addressing the limitations of sparse training. To better facilitate the combination of the two components, we further propose a hierarchical sparse strategy. The dense down-projection layer provides robust intermediate features for the model, while multiple sparse up-projection layers increase the capacity of the model. It's worth noting that our approach, as a \textit{general concept} of enhancing parameter utilization, can be applied to various adapter structures and other addition-based PEFT methods. 
On 27 diverse downstream visual tasks, our MoSA consistently outperforms all other fine-tuning methods, including full fine-tuning. Compared to full fine-tuning, MoSA achieves a lead of 1.36\% (on FGVC), 2.43\% (on VTAB) and 1.51\% (on GICD), while updating only around 1\% of the backbone parameters. Compared to the standard adapter, MoSA achieves an improvement of 1.32\% (on FGVC) and 1.06\% (on VTAB) without adding any computational or storage overhead. Additionally, we conducted comprehensive ablation experiments to validate the effectiveness of each component in our design. The results demonstrate that MoSA is indeed an Adapter Tuning method that successfully balances efficiency and performance.

We summarize our main contributions as follows:
\begin{itemize}
    \item We propose a novel Adapter Tuning method, namely \textbf{MoSA}, for fully unleashing the potential of the standard adapters. Through a mixed and sparse training approach involving splitting and merging, our method maximizes parameter efficiency, enhancing the performance of adapters in visual tasks.
    
    \item MoSA best achieves a balance between efficiency and performance. It exhibits efficiency in all the sparsification, mixed training, and inference stages, and the mutual promotion between stochastic activation and sparse training further enhances performance.
    
    \item We evaluate our method on a total of 27 downstream visual tasks spanning different domains, and MoSA significantly outperforms full fine-tuning as well as all other PEFT baselines, demonstrating the rationality of our design.
    
    \item We conduct comprehensive ablation studies to explore various design choices, demonstrating the effectiveness of each component. Additionally, we showcase the consistent improvements brought by MoSA across multiple model scales, architectures, and different PEFT methods.
\end{itemize}

\section{Related Work}
\label{sec:related}

\noindent \textbf{Parameter-efficient fine-tuning (PEFT).}
Recently, many large-scale pre-trained models (\eg LLaMA2~\cite{touvron2023llama2} 70B, GPT-3~\cite{brown2020gpt3} 175B) have emerged in deep learning research, which can achieve excellent performance in a variety of downstream tasks. However, updating and storing all model parameters for each task has become far more expensive. PEFT achieves efficiency in training and storage by updating only a small fraction of parameters compared to pre-trained models~\cite{zaken2021bitfit, li2021prefix, lester2021prompt, jia2022vpt, jie2023fact, lian2022ssf, liu2022ia3, liu2022noah, zhang2023gps, he2023spt}, among which Adapter Tuning~\cite{houlsby2019adapter, pfeiffer2020adapterfusion, pfeiffer2020adapterhub, chen2023samadapter, pan2022stadapter, park2023dualpath} is one of the most widely used. Due to space constraints, we focus exclusively on Adapter Tuning. For a broader overview of other PEFT methods, we recommend readers refer to surveys on tuning~\cite{ding2022deltatuning, yu2023visualtuning}. Adapters~\cite{houlsby2019adapter, pfeiffer2020adapterfusion} are first introduced in natural language processing (NLP), achieving efficient knowledge transfer by only updating the parameters of newly added lightweight bottleneck modules. AdaptFormer~\cite{chen2022adaptformer} first applies adapters to visual recognition, achieving remarkable results in video understanding. Subsequent works~\cite{hu2021lora, karimi2021compacter, he2022kadaptation} implement low-rank adaptation through various decomposition manners, further reducing the number of parameters required for fine-tuning. MoA~\cite{lee2023moa} addresses the domain generalization  through a MoE version of the aforementioned adaptation methods. Recently, SparseAdapter~\cite{he2022sparseadapter} enhances parameter efficiency through pruning of adapters before tuning. Although both MoE and sparsification can improve adapter performance, they also introduce additional computational costs. In this work, we propose a novel Adapter Tuning framework that enables two techniques to mutually enhance each other, ensuring both efficiency and performance.

\noindent \textbf{Mixture-of-Experts (MoE).}
The origin proposal of MoE~\cite{jacobs1991moe} is targeted at enhancing model capacity. The earliest proposed technologies, known as soft MoE~\cite{ma2018softmoe}, integrate outputs of multiple experts through weighted summation. However, this technique significantly increases the computational cost during training. To address this issue, the sparsely-gated MoE~\cite{shazeer2017sparsemoe, zhang2021moefication, he2023spt} was introduced, selecting specific experts for activation and directly assigning data to them, thereby reducing the computational burden. Nonetheless, this method often resulted in load imbalance, with some experts becoming inactive in later training stages, risking system collapse. THOR~\cite{zuo2021thor} effectively mitigates this issue by implementing random activation, which not only enhances efficiency but also improves overall model performance. AdaMix~\cite{wang2022adamix} extends this mechanism to efficient tuning, treating each PEFT module as an expert and achieving performance improvements in NLP tasks. However, the stochastic activation of multiple experts may lead to the issue of data dilution issue, resulting in suboptimal performance when there is insufficient data for downstream tasks. To address this problem, we employ sparse training for each expert, reducing the data requirement per expert. This approach significantly enhances performance while maintaining the same computational efficiency as standard adapters.

\noindent \textbf{Pruning and sparse training.}
The primary goal of model pruning~\cite{han2016eie, molchanov2017variational, lyu2019advances} aims to minimize deployment costs by reducing model parameters without significantly impacting its performance. Recent studies~\cite{guo2020diffpruning, xu2021childtuning, ansell2021sparsetuning} have revealed that pruning can enhance model fine-tuning. A reduced number of trainable parameters can act as an additional regularization constraint, potentially improving performance~\cite{zhang2023gps}. As a PEFT method, the architecture of adapters~\cite{houlsby2019adapter, pfeiffer2020adapterfusion} inherently possesses few updatable parameters. SparseAdapter~\cite{ansell2021sparsetuning} extends this concept by pre-pruning the adapter, enhancing parameter efficiency while maintaining or even improving performance compared to standard adapters. However, excessive sparsity can lead to training instability and suboptimal results on certain datasets. We take one step further to solve aforementioned weakness via integrating the MoE paradigm into our sparse tuning process. This integration not only achieves a sparse network structure, but also implements sparse activation during the training phase, which overcomes the inherent constraints of using a single sparse network and increases the model capacity.

\section{Method}
\label{sec:method}

\begin{figure*}[t]
  \centering
  \includegraphics[width=\linewidth]{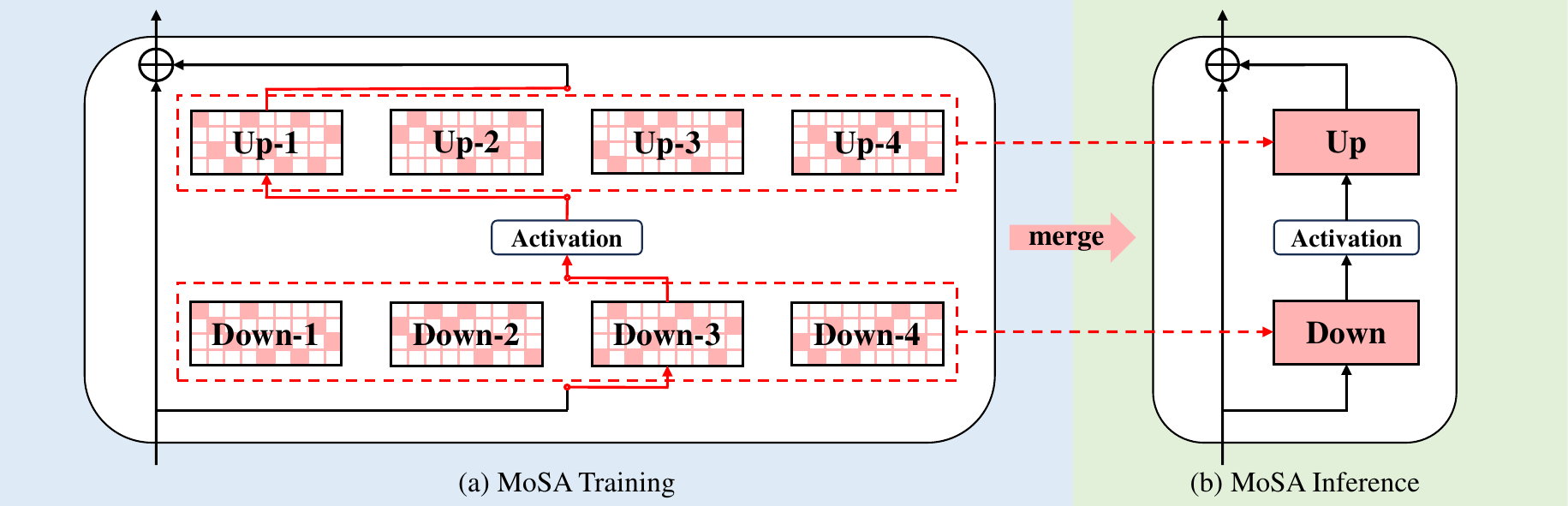}
  \caption{Architecture design of Mixture of Sparse Adapters. In the training phase, MoSA stochastically activates a sparse adapter during each forward pass; in the inference phase, MoSA merges multiple sparse adapters into a complete one to enhance efficiency.}
  \label{fig:architecture}
\end{figure*}

To achieve the optimal balance between efficiency and performance, we propose Mixture of Sparse Adapters (MoSA). The overall design of our method is shown in \Cref{fig:architecture}. We first overview our method in \Cref{sec:method-overview}. Subsequently, we describe how we split the standard adapter into multiple sparse adapters in \Cref{sec:method-split}. Then, we introduce the training strategy with stochastic activation of multiple sparse experts in \Cref{sec:method-stochastic}. Finally, we merge the split sparse adapters into a complete one for efficiency during inference in \Cref{sec:method-merge}.

\subsection{Overview}
\label{sec:method-overview}

Adapter Tuning is commonly employed in Transformer-based networks, which adapts the pre-trained representation to the target domain by injecting bottleneck modules into the Transformer layer. The general form of Adapter Tuning can be represented as:
\begin{equation}
  \widetilde{x} \leftarrow x + f(x \cdot \mathcal{W}^{\text{down}}) \cdot \mathcal{W}^{\text{up}}
  \label{eq:adapter}
\end{equation}
where $x \in \mathbb{R}^d$ represents the input of the adapter, $f$ is the activation function, $\mathcal{W}^{\text{down}} \in \mathbb{R}^{d \times r}$ and $\mathcal{W}^{\text{up}} \in \mathbb{R}^{r \times d}$ denote the linear layers for down-projection and up-projection, respectively. $r$ is the bottleneck dimension, satisfying $r \ll d$, which allows for a reduction in the number of parameters in the adapter. Increasing $r$ can enhance the performance of the adapter but also increases the total number of parameters. SparseAdapter~\cite{he2022sparseadapter} further exploits the high efficiency of parameters by pruning $\mathcal{W}^{\text{down}}$ and $\mathcal{W}^{\text{up}}$ before tuning. In our work, we split the standard adapter into $N$ sparse adapter experts $\mathscr{W}^{\text{down}} = \{\mathcal{W}_{i}^{\text{down}}\}_{i=1}^{N}, \mathscr{W}^{\text{up}} = \{\mathcal{W}_{i}^{\text{up}}\}_{i=1}^{N}$, and stochastically activate one of them during training for adaptation:
\begin{equation}
  \widetilde{x} \leftarrow x + f(x \cdot \mathcal{W}_{i}^{\text{down}}) \cdot \mathcal{W}_{j}^{\text{up}}
  \label{eq:stochastic_adapter}
\end{equation}
where $i, j \in \{1, \cdots, N\}$, representing the stochastic selection of $N$ experts.

\subsection{Sparse Adapter Splitting}
\label{sec:method-split}

To make the most of all the parameters within a single adapter, we first split the standard adapter into multiple sparse adapters, as illustrated in \Cref{fig:schema}(c). As all parameters in the adapter need to be updated, unlike \cite{he2022sparseadapter}, we do not adopt the task-specific pruning mechanism but instead employ a random splitting method to achieve as balanced grouping as possible, which also avoids the additional overhead of gradient computation based on downstream tasks before fine-tuning. 
Given the parameter of adapter $\mathcal{W} \in \mathbb{R}^{d \times r}$, we first generate an initial score $\mathcal{S} \sim \text{Uniform}(0, 1)  \in \mathbb{R}^{d \times r}$. Then, we split $\mathcal{W}$ into $N$ non-overlapping sparse adapters $\mathcal{W}_i: i \in \{1, \cdots, N\}$ by $\mathcal{S}$. Denoting the $i$-th $N$-quantile of all values in the matrix $\mathcal{S}$ as $s_i: i \in \{1, \cdots, N\}$, we obtain $N$ sparse masks $\mathcal{M}_i$ as follows:
\begin{equation}
  \mathcal{M}_i \leftarrow \mathbb{I} [s_{i-1} \leq \mathcal{S} < s_i]
  \label{eq:split}
\end{equation}
where $\mathbb{I}$ represents an all-ones matrix, and $s_0 = 0$. 

Once the masks are calculated, sparse gradient updating is performed according to the following form:
\begin{equation}
  \mathcal{W}_i' \leftarrow \mathcal{W}_i + \varepsilon \nabla \mathcal{L}(\mathcal{W}_i) \odot \mathcal{M}_i
  \label{eq:mask_tuning}
\end{equation}
where $\varepsilon$ represents the update step size, and $\nabla \mathcal{L}(\mathcal{W}_i)$ denotes the gradient of the task loss with respect to $\mathcal{W}_i$. In this way, the positions with a mask value of 0 have their gradients filtered to 0, thereby freezing the corresponding parameters and achieving sparse training for the adapter. Consequently, we obtain $N$ independent sparse adapters.

\subsection{Stochastic Activation Tuning}
\label{sec:method-stochastic}


With multiple sparse adapters, each can be treated as an expert, allowing for mixed training. Traditional MoE methods~\cite{shazeer2017sparsemoe, zhang2021moefication} involve a routing mechanism, introducing additional computation and load imbalance. Considering that Adapter Tuning itself serves as a parameter-efficient method, and inspired by ~\cite{zuo2021thor, wang2022adamix}, in the training process of MoSA, we also adopt a completely stochastic activation mechanism, which has been demonstrated to be simple yet effective in the following experiments. Given two partitioned parameter sets $\mathscr{W}^{\text{down}}$ and $\mathscr{W}^{\text{up}}$, for each batch of data, we randomly sample one module $\mathcal{W}_i^{\text{down}} \in \mathscr{W}^{\text{down}}, \mathcal{W}_j^{\text{up}} \in \mathscr{W}^{\text{up}}$ from each set to form an adapter $\mathcal{W} = \{\mathcal{W}_i^{\text{down}}, \mathcal{W}_j^{\text{up}}\}$, as shown in \Cref{eq:stochastic_adapter}. This activation mechanism ensures consistency in parameters and computational load with standard Adapter Tuning during training. Additionally, activating different sparse modules each time enables the model to learn different representations, thereby increasing the model capacity.

\noindent \textbf{Hierarchical sparse strategy.}
Although the MoE system enhances the model capacity, the sparse activation also reduces the amount of data seen by individual experts, resulting in suboptimal performance, especially when training data for downstream tasks is limited. The pre-pruning and sparse training methods can mitigate this data dilution issue to a certain extent. Building upon this, we further propose a hierarchical sparse strategy. The adapter module consists of two parts: the down-projection layer and the up-projection layer. We keep the down-projection layer as a dense matrix, \ie $\mathscr{W}^{\text{down}} = \mathcal{W}^{\text{down}}$, and only introduce sparsity in the up-projection layer, \ie $\mathscr{W}^{\text{up}} = \{\mathcal{W}_{i}^{\text{up}}\}_{i=1}^{N}$. This adaptation process can be expressed as:
\begin{equation}
  \widetilde{x} \leftarrow x + f(x \cdot \mathcal{W}^{\text{down}}) \cdot \mathcal{W}_{i}^{\text{up}}
  \label{eq:hierarchical_sparse}
\end{equation}
The dense down-projection layer provides robust intermediate features by receiving all training data, while multiple sparse up-projection layers map features to different subspaces, enhancing the performance on downstream tasks. Ablation experiments demonstrate the effectiveness of this hierarchical sparse strategy.

\noindent \textbf{Deep feature alignment.}
Like \cite{zuo2021thor} and \cite{wang2022adamix}, we also apply consistency regularization to ensure that different experts provide similar results for the same task. However, considering that different sparse adapters need to be merged after tuning, we further propose a deep feature alignment strategy to ensure that model parameters are not incompatible when merged. Given that deep features in neural networks are generally universal, while shallow features are typically task-specific, to strike a balance between model capacity and consistency, we align only the deep features of the model. Specifically, for a neural network with $L$ layers, we align the features extracted by different experts in the first $L/2$ layers. The overall optimization objective is formulated as follows:
\begin{equation}
    \mathcal{L}(x, y) = \text{CE}(p_1, y) + \frac{\alpha}{2}\left(\text{KL}(p_1\|p_2) + \text{KL}(p_2\|p_1)\right) + \beta\sum_{i=1}^{L/2}\text{MSE}(f_1^i, f_2^i)
    \label{eq:optimization_object}
\end{equation}
where $p_1, p_2$ represent the predicted probabilities after two stochastic forward passes of $x$, and $f_i^i, f_2^i$ represent the intermediate features in the $i$-th layer. Here, CE is the cross-entropy loss, KL is the Kullback–Leibler divergence and MSE is the mean square error. $\alpha$ and $\beta$ are hyper-parameters that control the weight of regularization terms, which are simply set to $1.0$ in our main experiments.

\subsection{Jigsaw-like Adapter Merging}
\label{sec:method-merge}

During inference, we merge the trained multiple sparse adapters like puzzle pieces into a complete adapter, as illustrated in \Cref{fig:architecture}. The merging process can be expressed as follows:
\begin{equation}
  \overline{\mathcal{W}}[\mathcal{M}_i > 0] \leftarrow \mathcal{W}_i[\mathcal{M}_i > 0]
  \label{eq:merge}
\end{equation}
Here, $\mathcal{M}_i$ represents the sparse mask for $\mathcal{W}_i$ from \Cref{eq:mask_tuning}, and $\overline{\mathcal{W}}$ represents the merged complete projection layer weight. After merging, the inference phase can be represented as:
\begin{equation}
  \widetilde{x} \leftarrow x + f(x \cdot \overline{\mathcal{W}^{\text{down}})} \cdot \overline{\mathcal{W}^{\text{up}}}
  \label{eq:inference}
\end{equation}
Experiments demonstrate that the merged adapter outperforms stochastic activation during inference.

\section{Experiments}
\label{sec:exp}

We evaluate MoSA across a series of downstream recognition tasks, spanning various model scales, architectures, and PEFT methods. We first describe our experimental setup in \Cref{sec:exp-setup}. Subsequently, we present the main experimental results in \Cref{sec:exp-main}. Furthermore, we demonstrate the performance of MoSA applied to different backbone scales and PEFT methods in \Cref{sec:exp-more}. In addition, we conduct extensive ablation experiments in \Cref{sec:exp-ablation} to verify the effectiveness of each component in our design. Finally, we provide t-SNE visualization results in \Cref{sec:exp-visualization}.

\subsection{Experimental Setup}
\label{sec:exp-setup}
\textbf{Pre-trained backbones.} We experiment with two Transformer architectures in vision: ViT~\cite{dosovitskiy2020vit} and Swin Transformer~\cite{liu2021swin}. In our experiments, all models are pre-trained on ImageNet-21k~\cite{russakovsky2015imagenet}. We adhere to the original configurations of these models, such as the number of image patches divided and the inclusion of the [CLS] token, etc.  More details can be found in Appendix.

\noindent \textbf{Baselines.} We compare our MoSA with other Adapter Tuning methods and commonly used fine-tuning strategies:
\begin{itemize}
    \item \textbf{Full fine-tuning}: fully update the whole backbone. 
    \item \textbf{Linear probing}: fix the model backbone and only update the classifier.
    \item \textbf{BitFit (Bias)~\cite{zaken2021bitfit}}: fine-tune all the bias terms in the pre-trained backbone.
    \item \textbf{Visual prompt tuning (VPT)~\cite{jia2022vpt}}: add learnable embeddings as prompts to modify the input, in two versions: shallow (insert prompts only into the first layer) and deep (introduce prompts at every layer).
    \item \textbf{AdaptFormer~\cite{chen2022adaptformer}}: insert bottleneck modules with residual connections to the feed-forward network (FFN) of each Transformer layer.
    \item \textbf{SparseAdapter~\cite{he2022sparseadapter}}: prune the standard adapter before tuning and update the remaining parameters via masks.
\end{itemize}

\noindent \textbf{Downstream tasks.}
We evaluate MoSA against other baselines on the following three collections of datasets:
\begin{itemize}
    \item \textbf{FGVC}: This benchmark consists of 5 Fine-Grained Visual Classification tasks, including CUB-200-2011~\cite{wah2011cub}, NABirds~\cite{van2015nabirds}, Oxford Flowers~\cite{nilsback2008flowers}, Stanford Dogs~\cite{dataset2011dogs}, and Stanford Cars~\cite{gebru2017cars}, which are representative examples of this category. We directly adapt the public split for training and validation sets if available, or we just follow the splits in ~\cite{jia2022vpt}.
    \item \textbf{VTAB}: VTAB-1k~\cite{zhai2019vtab} benchmark consists of 19 visual classification tasks from 3 diverse domains: \textit{Natural}, \textit{Specialized} and \textit{Structured}. Each task contains only 1000 training examples, but potentially spanning up to 397 classes, poses a significant challenge.
    \item \textbf{GICD}: We also collect a benchmark of 3 General Image Classification Datasets, including CIFAR-100~\cite{krizhevsky2009cifar100}, Aircraft~\cite{maji2013aircraft} and Food-101~\cite{bossard2014food101}, to demonstrate the efficiency of MoSA. All the datasets comprise around 100 categories, each containing at least 10,000 images, all of which are common objects in natural scenes. 
\end{itemize}

\noindent \textbf{Implementation details.} For all datasets, we only process the images with a randomly resized crop
to $224 \times 224$ and a random horizontal flip for data augmentation, instead of other strong augmentation and regularization strategies, like mixup~\cite{zhang2017mixup} and cutmix~\cite{yun2019cutmix}. We adopt the AdamW~\cite{loshchilov2017adamw} optimizer to fine-tune the pre-trained model for 100 epochs, with a linear warm-up of the learning rate for the first 10 epochs. For a fair comparison, we set the general hyperparameters to be the same in all Adaptor Tuning methods, including our MoSA. All experiments are conducted using the PyTorch~\cite{paszke2019pytorch} library on NVIDIA V100 and A100 GPUs. More implementation details can be found in Appendix.

\subsection{Main Results}
\label{sec:exp-main}
We provide a comprehensive evaluation of the effectiveness of our MoSA by comparing it with other baselines across 3 sets of up to 27 different datasets. In the following experiments, Top-1 accuracy (\%) is used to evaluate the performance of the methods on the respective datasets, and the number (M) of extra parameters (trainable parts excluding the classifier) is used to assess the efficiency of the methods. The best accuracy is highlighted in \textbf{bold}, while the second one is \underline{underlined}. The results of our method is highlighted with a \colorbox{pink}{red} background.

\begin{table}[t]
    \centering
    \caption{Results on FGVC with ViT-B/16 backbone pre-trained on ImageNet-21K}
    \label{tab:fgvc}
    \resizebox{\linewidth}{!}{
        \begin{tabular}{c|ccccc|cc}
        \toprule
        \diagbox{Method}{Dataset} & \makecell[c]{CUB-200\\-2011} & NABrids & \makecell[c]{Oxford\\Flowers} & \makecell[c]{Stanford\\Dogs} & \makecell[c]{Stanford\\Cars} & \makecell[c]{Mean\\Acc. (\%)} & \makecell[c]{Mean\\Params. (M)} \\
        \midrule
        Full fine-tuning & 87.3 & 82.7 & 98.8 & 89.4 & \textbf{84.5} & 88.54 & 85.80 \\
        \midrule
        Linear probing & 85.3 & 75.9 & 97.9 & 86.2 & 51.3 & 79.32 & 0.00 \\
        BitFit~\cite{zaken2021bitfit} & 88.4 & 84.2 & 98.8 & 91.2 & 79.4 & 88.40 & 0.10 \\
        \midrule
        VPT-shallow~\cite{jia2022vpt} & 86.7 & 78.8 & 98.4 & 90.7 & 68.7 & 84.62 & 0.27 \\
        VPT-deep~\cite{jia2022vpt} & \underline{88.5} & 84.2 & \underline{99.0} & 90.2 & \underline{83.6} & \underline{89.11} & 0.84 \\
        \midrule
        AdaptFormer~\cite{chen2022adaptformer} & 87.4 & 84.8 & \underline{99.0} & 90.7 & 81.0 & 88.58 & 1.54 \\
        SparseAdapter~\cite{he2022sparseadapter} & 87.8 & \underline{85.1} & 98.9 & \underline{91.4} & 80.3 & 88.70 & 0.39 \\
        \rowcolor{pink} MoSA (Ours) & \textbf{89.3} & \textbf{85.7} & \textbf{99.2} & \textbf{91.9} & 83.4 & \textbf{89.90} & 1.54  \\
        \bottomrule
        \end{tabular}
    }
\end{table}

\noindent \textbf{Fine-grained classification performance.} 
We first evaluate the effectiveness of our method on 5 widely used fine-grained visual classification tasks with ViT-B/16~\cite{dosovitskiy2020vit} backbone. As shown in \Cref{tab:fgvc}, our MoSA beats other baselines, including full fine-tuning, by a significant margin. Across the 5 downstream tasks, MoSA achieves an average accuracy of 89.90\%, surpassing full fine-tuning and AdaptFormer~\cite{chen2022adaptformer} by 1.36\% and 1.32\%, respectively, while maintaining the same number of trainable parameters as \cite{chen2022adaptformer}. Interestingly, SparseAdapter~\cite{he2022sparseadapter}, with fewer trainable parameters just through pruning, outperforms the standard adapter by 0.12\% on average, demonstrating the parameter redundancy in adapters for visual tasks. However, the high sparsity level results in ineffective utilization of most parameters in the adapter, limiting the overall performance gain. MoSA increases model capacity by mixed training of multiple sparse adapters, achieving an additional improvement of 1.20\%. Experiments show that our method enhances the performance of existing methods without introducing extra parameters or computation, maximizing the potential of adapters.

\begin{table}[t]
    \centering
    \caption{Results on VTAB with ViT-B/16 backbone pre-trained on ImageNet-21K}
    \label{tab:vtab}
    \begin{tabular}{c|ccc|cc}
    \toprule
    \diagbox{Method}{Dataset} & Natural & Specialized & Structured & \makecell[c]{Mean\\Acc. (\%)} & \makecell[c]{Mean\\Params. (M)} \\
    \midrule
    Full fine-tuning & 75.88 & 83.36 & 47.64 & 68.96 & 85.80 \\
    \midrule
    Linear probing & 68.93 & 77.16 & 26.84 & 57.64 & 0.00 \\
    BitFit~\cite{zaken2021bitfit} &  73.30 & 78.25 & 44.09 & 65.21 & 0.10 \\
    \midrule
    VPT-shallow~\cite{jia2022vpt} & 76.81 & 79.66 & 46.98 & 64.85 & 0.11 \\
    VPT-deep~\cite{jia2022vpt} & \underline{78.48} & 82.43 & \textbf{54.98} & 69.42 & 0.98 \\
    \midrule
    AdaptFormer~\cite{chen2022adaptformer} & 78.42 & \underline{83.41} & 49.17 & \underline{70.33} & 0.30 \\
    SparseAdapter~\cite{he2022sparseadapter} & 77.58 & 81.99 & 48.26 & 69.28 & 0.08 \\
    \rowcolor{pink} MoSA (Ours) & \textbf{79.86} & \textbf{84.03} & \underline{50.28} & \textbf{71.39} & 0.30 \\
    \bottomrule
    \end{tabular}
\end{table}

\noindent \textbf{Low-resource visual adaptation performance.} 
We also compare our method with other fine-tuning approaches on VTAB, which contains 19 diverse downstream tasks with only 1000 training samples per task, making it extremely challenging. Previous stochastically activated MoE methods, like ~\cite{zuo2021thor}, suffer from severe performance degradation when training data is insufficient. However, our MoSA overcomes this issue with its sparse training paradigm and hierarchical sparse strategy. The results in \Cref{tab:vtab} demonstrate that MoSA outperforms all other baselines by updating only 0.35\% (0.30M in 85.80M) of the pre-trained backbone parameters. Specifically, across 3 domains in VTAB, MoSA surpasses full fine-tuning by 3.98\%, 0.67\% and 2.64\%, while outperforming the second-best AdaptFormer by 1.44\%, 0.62\% and 1.11\%, providing strong evidence for the effectiveness of our design.

\begin{table}[t]
    \centering
    \caption{Results on GICD with ViT-B/16 backbone pre-trained on ImageNet-21K}
    \begin{tabular}{c|ccc|cc}
    \toprule
    \diagbox{Method}{Dataset} & CIFAR-100 & Aircraft & Food-101 & \makecell[c]{Mean\\Acc. (\%)} & \makecell[c]{Mean\\Params. (M)} \\
    \midrule
    Full fine-tuning & 89.12 & 70.93 & \underline{90.96} & 83.67 & 85.80 \\
    \midrule
    Linear probing & 85.95 & 45.06 & 88.14 & 73.05 & 0.00 \\
    BitFit~\cite{zaken2021bitfit} & 91.69 & 68.71 & 89.59 & 83.33 & 0.10 \\
    \midrule
    AdaptFormer~\cite{chen2022adaptformer} & \underline{91.86} & \underline{71.71} & 90.89 & \underline{84.82} & 1.19 \\
    SparseAdapter~\cite{he2022sparseadapter} & 91.20 & 67.15 & 89.37 & 82.57 & 0.30 \\
    \rowcolor{pink} MoSA (Ours) & \textbf{92.22} & \textbf{72.14} & \textbf{91.17} & \textbf{85.18} & 1.19 \\
    \bottomrule
    \end{tabular}
    \label{tab:gicd}
\end{table}

\noindent \textbf{General large-scale classification performance.} 
To further evaluate the generality of our method, we compare MoSA with other fine-tuning methods on 3 general classification tasks, as shown in \Cref{tab:gicd}. MoSA outperforms all baselines, including full fine-tuning, on all datasets. Specifically, compared to AdaptFormer and full fine-tuning, MoSA achieves an average accuracy improvement of 1.51\% and 0.36\%, respectively, without introducing any additional parameters. It is worth noting that SparseAdapter performs poorly on this benchmark, exhibiting an accuracy drop of 2.25\% compared to the standard adapter. This is attributed to the fact that the three datasets in GICD contain relatively sufficient training data, mitigating the advantages of sparsity. However, our method, through multiple sparse adapter experts, demonstrates robust performance in scenarios with both abundant and limited training data, outperforming SparseAdapter by 2.61\%, which proves the soundness of our design.



\subsection{Extended Results on Different Backbone Scales and LoRA}
\label{sec:exp-more}
In this section, we validate the performance of MoSA on different backbone scales and PEFT methods. More experiments on different model architectures (\eg Swin Transformer) and adapter structures can be found in Appendix. 

\begin{table}[t]
    \centering
    \caption{Results on FGVC with ViT-L/16 backbone pre-trained on ImageNet-21K}
    \vspace{-3mm}
    \label{tab:vit-l}
    \resizebox{\linewidth}{!}{
        \begin{tabular}{c|ccccc|cc}
        \toprule
        \diagbox{Method}{Dataset} & \makecell[c]{CUB-200\\-2011} & NABrids & \makecell[c]{Oxford\\Flowers} & \makecell[c]{Stanford\\Dogs} & \makecell[c]{Stanford\\Cars} & \makecell[c]{Mean\\Acc. (\%)} & \makecell[c]{Mean\\Params. (M)} \\
        \midrule
        Full fine-tuning & 88.3 & 85.9 & 96.7 & 93.1 & \textbf{86.8} & 90.16 & 303.30 \\
        \midrule
        Linear probing & 84.7 & 78.9 & 97.4 & 89.6 & 55.1 & 81.14 & 0.00 \\
        BitFit~\cite{zaken2021bitfit} & 88.5 & 86.4 & 98.8 & 93.5 & 83.2 & 90.08 & 0.27 \\
        \midrule
        AdaptFormer-64~\cite{chen2022adaptformer} & 89.3 & 86.3 & 98.8 & 93.2 & 80.9 & 89.70 & 3.17 \\
        SparseAdapter-64~\cite{he2022sparseadapter} & \underline{89.4} & \underline{86.8} & \underline{99.0} & \underline{93.9} & 82.3 & \underline{90.28} & 0.79 \\
        \rowcolor{pink} MoSA-64 (Ours) & \textbf{89.7} & \textbf{87.2} & \textbf{99.4} & \textbf{94.5} & \underline{84.9} & \textbf{91.06} & 3.17  \\
        \bottomrule
        \end{tabular}
    }
    \vspace{-3mm}
\end{table}

\noindent \textbf{MoSA on different backbone scales.} 
Here we evaluate MoSA with a larger backbone ViT-L/16 (303.3M \vs 85.8M ViT-B/16). The results in \Cref{tab:vit-l} indicate that, with the increase in backbone scale, the performance of the standard adapter cannot surpass full fine-tuning (89.70\% \vs 90.16\%). Sparse training could improve the performance of adapters , leading by 0.58\% and 0.12\% over the standard adapter and full fine-tuning, respectively. Our MoSA further enhances the performance by 0.78\%, outperforming full fine-tuning by a large margin. This experiment thoroughly demonstrates the importance of sparse training as the scale of pre-trained models increases.

\begin{table}[t]
    \centering
    \caption{Results on FGVC with LoRA}
    \vspace{-3mm}
    \label{tab:lora}
    \resizebox{\linewidth}{!}{
        \begin{tabular}{c|ccccc|cc}
        \toprule
        \diagbox{Method}{Dataset} & \makecell[c]{CUB-200\\-2011} & NABrids & \makecell[c]{Oxford\\Flowers} & \makecell[c]{Stanford\\Dogs} & \makecell[c]{Stanford\\Cars} & \makecell[c]{Mean\\Acc. (\%)} & \makecell[c]{Mean\\Params. (M)} \\
        \midrule
        LoRA-16~\cite{hu2021lora} & 87.2 & 83.5 & 98.6 & 89.3 & 83.7 & 88.44 & 0.59 \\
        SparseLoRA-16~\cite{he2022sparseadapter} & 87.4 & 84.9 & 98.9 & 91.1 & 79.9 & 88.44 & 0.15 \\
        \rowcolor{pink} MoSL-16 (Ours) & \textbf{89.0} & \textbf{85.6} & \textbf{99.3} & \textbf{91.8} & \textbf{83.9} & \textbf{89.92} & 0.59  \\
        \bottomrule
        \end{tabular}
    }
    \vspace{-3mm}
\end{table}

\noindent \textbf{MoSA on different PEFT methods.} 
To demonstrate the generality of our approach, we choose another widely used PEFT method, namely LoRA~\cite{hu2021lora}. LoRA achieves parameter-efficient fine-tuning by applying a low-rank decomposition to the weight updates of linear layers. During inference, the additional modules could be merged into the pre-trained model parameters, resulting in zero extra overhead during inference. In \Cref{tab:lora}, we compare the LoRA version sparse adapter (namely SparseLoRA) and our MoSA (namely MoSL) with the standard LoRA, with the rank of all LoRA modules set to 16. Our MoSL outperforms both the standard and sparse versions of LoRA, achieving an improvement of 1.48\%, further confirming the effectiveness and rationality of our design.

\subsection{Ablation Study}
\label{sec:exp-ablation}
We conduct comprehensive ablation studies to verify the effectiveness of each component in our MoSA design. All ablation experiments are performed on FGVC with ViT-B backbone, and the performances of different strategies are measured using mean accuracy over the datasets. The best component choice, which is also used in the main results, is highlighted with a \colorbox{lime}{green} background.

\begin{table}[t]
    \centering
    \begin{minipage}[t]{0.48\textwidth}
        \makeatletter\def\@captype{table}
        \caption{Hierarchical sparse strategy}
        \label{tab:ablation-hierarchical}
        \begin{tabular}{cc}
        \toprule
        Strategy & Mean Acc. (\%) \\
        \midrule
        dense down + dense up & 88.58 \\
        sparse down + sparse up & 88.91 \\
        \rowcolor{lime} dense down + sparse up & \textbf{89.90} \\
        \bottomrule
        \end{tabular}
    \end{minipage}
    \begin{minipage}[t]{0.48\textwidth}
        \makeatletter\def\@captype{table}
        \caption{Consistency regularization}
        \label{tab:ablation-regularization}
        \begin{tabular}{cc}
        \toprule
        Regularization & Mean Acc. (\%) \\
        \midrule
        none & 88.93 \\
        consistency & 89.25 \\
        \rowcolor{lime} consistency + alignment & \textbf{89.90} \\
        \bottomrule
        \end{tabular}
    \end{minipage}
\end{table}

\noindent \textbf{Hierarchical sparse strategy.}
The MoE system increases the model capacity, while the sparse gating mechanism also leads to a decrease in the amount of data each expert encounters, resulting in suboptimal performance particularly when training data for downstream tasks is limited. In order to address this issue, we propose a hierarchical sparse strategy, and the results in \Cref{tab:ablation-hierarchical} demonstrate the effectiveness of this design. Applying the sparse strategy to both the down-projection and up-projection layer only achieves an accuracy of 88.91\%, slightly (0.33\%) ahead of Adaptformer (dense down-projection and up-projection layer). However, with the hierarchical sparse strategy, we preserve the down-projection layer as a dense matrix while sparsely splitting the up-projection layer. In this way, MoSA outperforms Adaptformer by a large margin (89.90\% \vs 88.58\%), demonstrating the importance of this hierarchical strategy.

\begin{table}[t]
    \centering
    \begin{minipage}[t]{0.53\textwidth}
        \makeatletter\def\@captype{table}
        \caption{Different alignment strategies}
        \label{tab:ablation-alignment}
        \resizebox{\linewidth}{!}{
            \begin{tabular}{c|cccc}
            \toprule
            Alignment position     & none  & shallow & deep  & all   \\
            \midrule
            Mean Acc. (\%) & 89.25 & \textbf{89.90}   & 88.84 & 88.79 \\
            \bottomrule
            \end{tabular}
        }
    \end{minipage}
    \begin{minipage}[t]{0.43\textwidth}
        \makeatletter\def\@captype{table}
        \resizebox{0\linewidth}{!}{
            \begin{tabular}{ccc}
            \toprule
            Mechanism & FLOPs & Mean Acc. (\%) \\
            \midrule
            fixed & $1\times$ & 88.42 \\
            stochastic & $1\times$ & 89.27 \\
            ensemble & $N\times$ & 88.63 \\
            \rowcolor{lime} merge & $1\times$ & \textbf{89.90} \\
            \bottomrule
            \end{tabular}
        }
    \end{minipage}
    
    \begin{minipage}[t]{0.53\textwidth}
        \makeatletter\def\@captype{table}
        \caption{Impact of expert number}
        \label{tab:ablation-expert}
        \resizebox{\linewidth}{!}{
            \begin{tabular}{c|cccccc}
            \toprule
            Expert number & 1 & 2 & 3 & 4 & 5 & 8 \\
            \midrule
            Mean Acc. (\%) & 88.58 & 89.59 & 89.42 & 89.43 & 89.05 & 88.57  \\
            \bottomrule
            \end{tabular}
        }
    \end{minipage}
    \begin{minipage}[t]{0.43\textwidth}
        \vspace{-18mm}
        \makeatletter\def\@captype{table}
        \caption{Different inference mechanisms and corresponding efficiency}
        \label{tab:ablation-inference}
        \resizebox{\linewidth}{!}{
            \begin{tabular}{ccc}
            \toprule
            Mechanism & FLOPs & Mean Acc. (\%) \\
            \midrule
            fixed & $1\times$ & 88.42 \\
            stochastic & $1\times$ & 89.27 \\
            ensemble & $N\times$ & 88.63 \\
            \rowcolor{lime} merge & $1\times$ & \textbf{89.90} \\
            \bottomrule
            \end{tabular}
        }
    \end{minipage}
    \vspace{-3mm}
\end{table}

\noindent \textbf{Consistency regularization and feature alignment.} The mechanism of stochastic activation may lead to significant discrepancies among different experts, which can have adverse effects on the sparse module merging. In \Cref{tab:ablation-regularization}, we investigate the impact of regularization constraints in the training of MoSA. Without any regularization, our method achieves an accuracy of 88.93\%. Applying consistency regularization on the final outputs of the model results in an improvement of 0.32\%. And deep alignment of features extracted by different experts can further improve the accuracy by 0.65\%. We also explore the impact of different feature alignment positions on model performance. As shown in \Cref{tab:ablation-alignment}, performing alignment solely at shallow layers (closer to the input) can lead to a 0.65\% improvement. In contrast, executing alignment at deep layers (closer to the output) results in a 0.41\% decrease, attributable to sub-module collapse.

\noindent \textbf{Expert number.} To evaluate the compatibility between sparse training and MoE, we vary the number of splits in the adapters of MoSA during training, ranging from 1 to 5 and 8. The results are presented in \Cref{tab:ablation-expert}. When the number of splits is set to 1, our method degrades to AdaptFormer. We can see that when the number of experts is between 2 and 5, indicating a sparsity level between 20\% and 50\% for the adapters, MoSA consistently achieves good performance. However, when the number of experts increases to 8, the performance experiences a noticeable decline due to excessive sparsity.

\noindent \textbf{Merging \vs ensembling.} During inference, we merge multiple sparse adapters to form a complete one. However, it has been pointed out that stochastic activation during inference can also achieve good performance~\cite{zuo2021thor}. So we compare various inference methods, including fixed activation, stochastic activation, logits ensembling, and parameter merging. Results are presented in \Cref{tab:ablation-inference}. It can be observed that fixed activation has the lowest accuracy, reaching only 88.41\% on average. In comparison, stochastic activation achieves an improvement of 0.85\%. Logits ensembling also leads to a certain improvement over fixed activation, but the increase of only 0.21\% comes at the cost of $N$ times computational complexity ($N$ refers to the expert number), significantly reducing the inference speed. Finally, our parameter merging method achieves the best performance without any additional computation, further enhancing 0.63\% over stochastic activation.

\begin{figure}[t]
    \centering
    \begin{minipage}[t]{0.31\textwidth}
        \includegraphics[width=\linewidth]{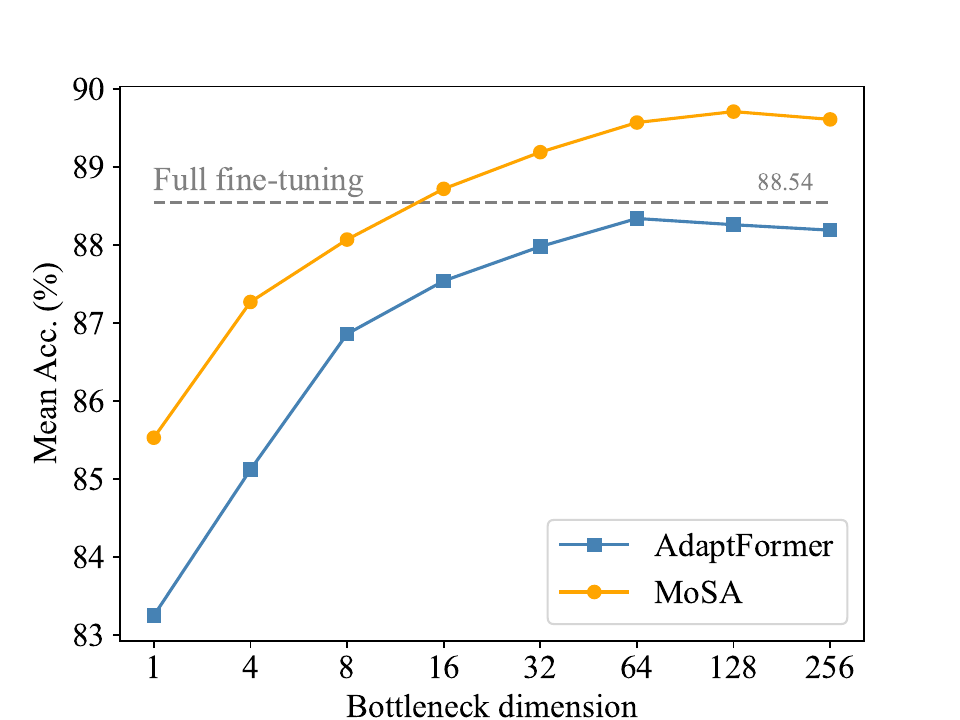}
        \caption{Impact of adapter bottleneck dimensions}
        \label{fig:ablation-bottleneck}
    \end{minipage}
    \begin{minipage}[t]{0.64\textwidth}
        \includegraphics[width=\linewidth]{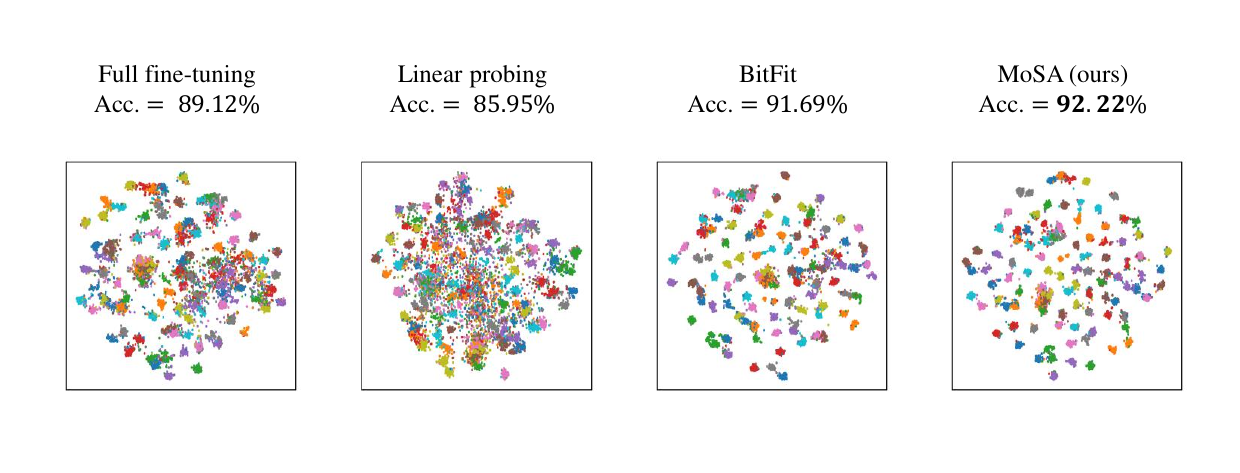}
        \caption{t-SNE visualization on CIFAR-100}
        \label{fig:visualization}
    \end{minipage}
    \vspace{-3mm}
\end{figure}

\noindent \textbf{Adapter bottleneck dimension.} \Cref{fig:ablation-bottleneck} shows the impact of bottleneck dimensions of the adapter in AdaptFormer and MoSA. Overall, the performance shows an increasing trend followed by a decline as the number of trainable parameters increases. Across all bottleneck dimensions, MoSA consistently outperforms AdaptFormer, and our method can even surpass full fine-tuning with a bottleneck dimension of only 16. It is worth noting that the performance of AdaptFormer starts to decline when the bottleneck dimension exceeds 64, while in our method, this turning point occurs at 128. This indicates that our design can more effectively utilize all parameters within the adapter.

\subsection{Visualization}
\label{sec:exp-visualization}
Here, we provide t-SNE visualizations to show the feature distribution of different methods on the CIFAR-100 dataset. The results are shown in \Cref{fig:visualization}. We can observe that our MoSA achieves better feature clustering compared to full fine-tuning and other PEFT baselines.

\section{Conclusion}
\label{sec:conclusion}

In this paper, we focus on the Adapter Tuning method in parameter-efficient fine-tuning and propose MoSA to improve the performance of standard adapters without any extra parameters or computation. Recognizing that standard adapters still suffer from parameter redundancy, we combine sparse training with multiple stochastically activated experts to fully utilize all parameters within the adapters. Comprehensive experiments on a total of 27 datasets show that MoSA consistently outperforms all other baselines, achieving state-of-the-art performance in Adapter Tuning. We hope that our work could inspire researchers to reconsider the issue of parameter redundancy in adapters and make further advancements towards more efficient PEFT methods.

\clearpage
\setcounter{section}{0}
\renewcommand\thesection{\Alph{section}}

\section{Dataset Details}
\label{sec:dataset}

Here, we describe the details of all datasets used to validate MoSA. The number of classes and the train/valid/test splits for each dataset are shown in \Cref{tab:dataset}.
\begin{itemize}
\item \textbf{FGVC}: Fine-Grained Visual Classification (FGVC) benchmark consists of 5 downstream tasks, which are CUB-200-2011~\cite{wah2011cub}, NABirds~\cite{van2015nabirds}, Oxford Flowers~\cite{nilsback2008flowers}, Stanford Dogs~\cite{dataset2011dogs} and Stanford Cars~\cite{gebru2017cars}. Each task contains more than 100 classes and a few thousand images.
\item \textbf{VTAB}: Visual Task Adaptation Benchmark~\cite{zhai2019vtab} (VTAB) contains 19 visual classification tasks grouped into 3 domains: (1) \textit{Natural} - tasks with natural images captured by standard cameras; (2) \textit{Specialized} - tasks with images captured via specialized equipment, \eg medical camera and satellite sensor; (3) \textit{Structured} - tasks with images synthesized from simulated environments, which require geometric comprehension like object counting and depth estimation. Each task of VTAB contains only 1000 training samples, but may span up to 397 classes with several thousand testing samples, making it highly challenging.
\item \textbf{GICD}: General Image Classification Datasets (GICD) benchmark consists of 3 general classification tasks, which are CIFAR-100~\cite{krizhevsky2009cifar100}, Aircraft~\cite{maji2013aircraft} and Food-101~\cite{bossard2014food101}. All the tasks comprise around 100 classes, each containing at least 10,000 images, all of which are common objects in natural scenes. 
\end{itemize}

\section{Implementation Details}
\label{sec:implementation}

In \Cref{tab:implementation}, we summarize all experimental configurations with Adapter Tuning and LoRA. For other baselines, we just follow \cite{jia2022vpt}. Following the linear scaling rule~
\cite{chen2021empirical, goyal2017sgd, krizhevsky2014trick, he2022mae}, the learning rate is set as $base\_lr\times b / 256$, where $b$ is the batch size and $base\_lr$ is chosen from the range specified in \Cref{tab:implementation}.

\begin{table}
    \centering
    \caption{Details of all datasets used to validate MoSA.}
    \resizebox{0.95\linewidth}{!}{
    \label{tab:dataset}
    \begin{tabular}{llllll}
    \toprule
    \multicolumn{2}{l}{Dataset} & \#Classes & Train & Val & Test \\
    \midrule
    \multicolumn{6}{c}{Fine-Grained Visual Classification (FGVC)} \\
    \midrule
    CUB-200-2011~\cite{wah2011cub} & & 200 & 5,394 & 600 & 5,794 \\
    NABirds~\cite{van2015nabirds} & & 555 & 21,536 & 2,393 & 24,633 \\
    Oxford Flowers~\cite{nilsback2008flowers} & & 102 & 1,020 & 1,020 & 6,149 \\
    Stanford Dogs~\cite{dataset2011dogs} & & 120 & 10,800 & 1,200 & 8,580 \\
    Stanford Cars~\cite{gebru2017cars} & & 196 & 7,329 & 815 & 8,041 \\
    \midrule
    \multicolumn{6}{c}{Visual Task Adaptation Benchmark (VTAB)~\cite{zhai2019vtab}} \\
    \midrule
    \multirow{7}{*}{Natural} & CIFAR-100~\cite{krizhevsky2009cifar100} & 100 & \multirow{7}{*}{800/1000} & \multirow{7}{*}{200} & 10,000 \\
     & Caltech101~\cite{fei2006caltech101} & 102 & & & 6,084 \\
     & DTD ~\cite{cimpoi2014dtd} & 47 & & & 1,880 \\
     & Flowers102~\cite{nilsback2008flowers} & 102 & & & 6,149 \\
     & Pets ~\cite{parkhi2012pets} & 37 & & & 3,669 \\
     & SVHN~\cite{yuval2011svhn} & 10 & & & 26,032 \\
     & Sun397~\cite{xiao2010sun397} & 397 & & & 21,750 \\
    \midrule
    \multirow{4}{*}{Specialized} & Patch Camelyon~\cite{veeling2018patch} & 2 & \multirow{4}{*}{800/1000} & \multirow{4}{*}{200} & 32,768 \\
     & EuroSAT~\cite{helber2019eurosat} & 10 & & & 5,400 \\
     & Resisc45~\cite{cheng2017resisc45} & 45 & & & 6,300 \\
     & Retinopathy~\cite{graham2015retinopathy} & 5 & & & 42,670 \\
    \midrule
    \multirow{8}{*}{Structured} & Clevr/count~\cite{johnson2017clevr} & 8 & \multirow{8}{*}{800/1000} & \multirow{8}{*}{200} & 15,000 \\
     & Clevr/distance~\cite{johnson2017clevr} & 6 & & & 15,000 \\
     & DMLab ~\cite{beattie2016dmlab} & 6 & & & 22,735 \\
     & KITTI/distance~\cite{geiger2013kitti} & 4 & & & 711 \\
     & dSprites/location~\cite{matthey2017dsprites} & 16 & & & 73,728 \\
     & dSprites/orientation~\cite{matthey2017dsprites} & 16 & & & 73,728 \\
     & SmallNORB/azimuth~\cite{lecun2004smallnorb} & 18 & & & 12,150 \\
     & SmallNORB/elevation~\cite{lecun2004smallnorb} & 9 & & & 12,150 \\
    \midrule
    \multicolumn{6}{c}{General Image Classification Datasets (GICD)} \\
    \midrule
    CIFAR-100~\cite{krizhevsky2009cifar100} & & 100 & 50,000 & - & 10,000 \\
    Aircraft~\cite{maji2013aircraft} & & 100 & 3,334 & 3,333 & 3,333 \\
    Food-101~\cite{bossard2014food101} & & 101 & 75,750 & - & 25,250 \\
    \bottomrule
    \end{tabular}
    }
\end{table}

\begin{table}
    \centering
    \caption{Implementation details for Adapter Tuning and LoRA.}
    \label{tab:implementation}
    \begin{tabular}{ll}
    \toprule
    Configuration & Value \\
    \midrule
    Optimizer & AdamW~\cite{loshchilov2017adamw} \\
    Base learning rate range & \{0.01, 0.005, 0.001, 0.0005, 0.0001\} \\
    Weight decay range & \{0.01, 0.0\} \\
    Learning rate schedule & cosine decay~\cite{loshchilov2016cosine} \\
    Batch size & 128 (ViT-B/16, Swin-B), 64 (ViT-L/16) \\
    Warmup epoch & 10 \\
    Total epoch & 100 (ViT-B/16, Swin-B), 50 (ViT-L/16) \\
    Augmentation & RandomResizedCrop~\cite{he2022mae}, RandomHorizontalFlip \\
    \bottomrule
    \end{tabular}
\end{table}

\section{More Experimental Results}
\label{sec:more_exp}

In this section, we validate MoSA on different backbone architectures and adapter structures. We also show the pre-task results for MoSA with ViT-B/16 on VTAB-1k. Similar to the main text, Top-1 accuracy (\%) is used to evaluate the performance of the methods on the respective datasets, and the number (M) of extra parameters (trainable parts excluding the classifier) is used to assess the efficiency. The best accuracy is highlighted in \textbf{bold}, while the second one is \underline{underlined}. The results of our method is highlighted with a \colorbox{pink}{red} background.
\vspace{3mm}

\begin{table}[t]
    \centering
    \caption{Results on FGVC with Swin-B backbone pre-trained on ImageNet-21K.}
    \label{tab:swin}
    \resizebox{\linewidth}{!}{
        \begin{tabular}{c|ccccc|cc}
        \toprule
        \diagbox{Method}{Dataset} & \makecell[c]{CUB-200\\-2011} & NABrids & \makecell[c]{Oxford\\Flowers} & \makecell[c]{Stanford\\Dogs} & \makecell[c]{Stanford\\Cars} & \makecell[c]{Mean\\Acc. (\%)} & \makecell[c]{Mean\\Params. (M)} \\
        \midrule
        Full fine-tuning & 88.2 & \textbf{87.8} & 99.0 & 85.5 & \textbf{90.2} & \underline{90.14} & 86.74 \\
        \midrule
        Linear probing & 87.8 & 83.8 & 98.8 & 84.7 & 69.2 & 84.86 & 0.00 \\
        BitFit~\cite{zaken2021bitfit} & 88.4 & 85.2 & 99.2 & 85.3 & 83.4 & 88.30 & 0.20 \\
        \midrule
        AdaptFormer-64~\cite{chen2022adaptformer} & \underline{89.7} & \underline{87.7} & 99.3 & 86.0 & \underline{87.7} & 90.08 & 1.55 \\
        SparseAdapter-64~\cite{he2022sparseadapter} & \underline{89.7} & 87.4 & \underline{99.4} & \underline{87.1} & 86.8 & 90.08 & 0.39 \\
        \rowcolor{pink} MoSA-64 (Ours) & \textbf{90.6} & \textbf{87.8} & \textbf{99.6} & \textbf{88.3} & 87.3 & \textbf{90.72} & 1.55 \\
        \bottomrule
        \end{tabular}
    }
\end{table}

\textbf{MoSA on different model architectures.} In addition to the standard ViT, we also experiment with another hierarchical vision Transformer, Swin-B~\cite{liu2021swin}, to demonstrate the effectiveness of MoSA. Similar to ViT, we can easily apply MoSA to Swin. As shown in \Cref{tab:swin}, due to the strong feature extraction capability of this model architecture, full fine-tuning performs well on Swin, while other PEFT methods show suboptimal performance. It's worth noting that our MoSA is the only method that outperforms full fine-tuning on Swin (90.72\% \vs 90.14\%), indicating that MoSA consistently adapts various vision Transformers to downstream tasks and improves performance.

\begin{table}[t]
    \centering
    \caption{Results on FGVC with different adapter structures.}
    \label{tab:adapter}
    \resizebox{\linewidth}{!}{
        \begin{tabular}{c|ccccc|cc}
        \toprule
        \diagbox{Method}{Dataset} & \makecell[c]{CUB-200\\-2011} & NABrids & \makecell[c]{Oxford\\Flowers} & \makecell[c]{Stanford\\Dogs} & \makecell[c]{Stanford\\Cars} & \makecell[c]{Mean\\Acc. (\%)} & \makecell[c]{Mean\\Params. (M)} \\
        \midrule
        Adapter-Pfeifferr~\cite{pfeiffer2020adapterfusion} & 84.5 & 81.3 & 97.9 & 88.8 & 76.7 & 85.84 & 1.21 \\
        SparseAdapter-Pfeiffer~\cite{he2022sparseadapter} & 86.7 & 83.9 & 98.5 & 90.0 & 77.6 & 87.34 & 0.30 \\
        \rowcolor{pink} MoSA-Pfeiffer (Ours) & \textbf{89.3} & \textbf{85.5} & \textbf{99.2} & \textbf{91.6} & \textbf{79.6} & \textbf{89.04} & 1.21  \\
        \midrule
        Adapter-Houlsby~\cite{houlsby2019adapter} & 87.5 & 81.9 & 97.9 & 89.0 & 75.7 & 86.40 & 2.38 \\
        SparseAdapter-Houlsby~\cite{he2022sparseadapter} & 87.5 & 83.3 & 98.9 & 90.1 & 78.7 & 87.70 & 0.59 \\
        \rowcolor{pink} MoSA-Houlsby (Ours) & \textbf{89.3} & \textbf{86.2} & \textbf{99.3} & \textbf{92.1} & \textbf{80.4} & \textbf{89.46} & 2.38  \\
        \bottomrule
        \end{tabular}
    }
\end{table}

\textbf{MoSA on different adapter structures.} As a supplement, we apply MoSA to two different adapter structures: Pfeifferr~\cite{pfeiffer2020adapterfusion} and Houlsby~\cite{houlsby2019adapter}. Both Pfeifferr and Houlsby use a sequential connection for adapter design, with Pfeifferr incorporating the adapters only after the FFN layers, while Houlsby includes the adapters after both the Attention and FFN layers. The performance comparison on the two adapter structures is shown in \Cref{tab:adapter}. In this experiment, the bottleneck dimension for all adapters is set to 64. It can be observed that on both structures, sparse training brings improvements of 1.50\% and 1.30\% over the standard adapter, and the corresponding versions of our MoSA further yield performance gains of 1.70\% and 1.76\%.

\begin{sidewaystable}[t]
    \centering
    \caption{Per-task results on VTAB-1k with ViT-B/16 pre-trained on ImageNet-21K.}
    \label{tab:vtab_detailed}
    \resizebox{\linewidth}{!}{
        \begin{tabular}{c|cccccccc|ccccc|ccccccccc}
        \toprule
        \multirow{2}{*}{\diagbox{Method \\ \\ \\}{\\ \\ \\ Dataset}} & \multicolumn{8}{c}{Natural} & \multicolumn{5}{c}{Specialized} & \multicolumn{9}{c}{Structured} \\
        \cline{2-23}
         & \rotatebox{90}{CIFAR-100} & \rotatebox{90}{Caltech101} & \rotatebox{90}{DTD} & \rotatebox{90}{Flowers102} & \rotatebox{90}{Pets} & \rotatebox{90}{SVHN} & \rotatebox{90}{Sun397} & \rotatebox{90}{Mean} & \rotatebox{90}{Patch Camelyon} & \rotatebox{90}{EuroSAT} & \rotatebox{90}{Resisc45} & \rotatebox{90}{Retinopathy} & \rotatebox{90}{Mean} & \rotatebox{90}{Clevr/count} & \rotatebox{90}{Clevr/distance} & \rotatebox{90}{DMLab} & \rotatebox{90}{KITTI/distance} & \rotatebox{90}{dSprites/loc} & \rotatebox{90}{dSprites/ori} & \rotatebox{90}{SmallNORB/azi} & \rotatebox{90}{SmallNORB/ele} & \rotatebox{90}{Mean} \\
        \midrule
        Full fine-tuning & 68.9 & 87.7 & 64.3 & 97.2 & 86.9 & \textbf{87.4} & 38.8 & 75.88 & 79.7 & \underline{95.7} & \textbf{84.2} & 73.9 & 83.36 & 56.3 & \underline{58.6} & \underline{41.7} & 65.5 & 57.5 & \underline{46.7} & 25.7 & 29.1 & 65.57 \\
        \midrule
        Linear probing & 63.4 & 85.0 & 64.3 & 97.0 & 86.3 & 36.6 & 51.0 & 68.93 & 78.5 & 87.5 & 68.6 & 74.0 & 77.16 & 34.3 & 30.6 & 33.2 & 55.4 & 12.5 & 20.0 & 9.6 & 19.2 & 53.00 \\
        BitFit~\cite{zaken2021bitfit} & 72.8 & 87.0 & 59.2 & 97.5 & 85.3 & 59.9 & 51.4 & 73.30 & 78.7 & 91.6 & 72.9 & 69.8 & 78.25 & 61.5 & 55.6 & 32.4 & 55.9 & 66.6 & 40.0 & 15.7 & 25.1 & 62.05 \\
        \midrule
        VPT-shallow~\cite{jia2022vpt} & 77.7 & 86.9 & 62.6 & 97.5 & 87.3 & 74.5 & 51.2 & 76.81 & 78.2 & 92.0 & 75.6 & 72.9 & 79.66 & 50.5 & \underline{58.6} & 40.5 & 67.1 & \underline{68.7} & 36.1 & 20.2 & \underline{34.1} & 64.85 \\
        VPT-deep~\cite{jia2022vpt} & 78.8 & \underline{90.8} & 65.8 & 98.0 & 88.3 & 78.1 & 49.6 & \underline{78.48} & \underline{81.8} & \textbf{96.1} & \underline{83.4} & 68.4 & 82.43 & 68.5 & \textbf{60.0} & \textbf{46.5} & \textbf{72.8} & \textbf{73.6} & \textbf{47.9} & \textbf{32.9} & \textbf{37.8} & \textbf{69.42} \\
        \midrule
        AdaptFormer~\cite{chen2022adaptformer} & 78.9 & 90.0 & \textbf{67.0} & \underline{98.7} & 89.0 & 72.2 & \underline{53.2} & 78.42 & 81.5 & \underline{95.7} & 81.2 & \textbf{75.2} & \underline{83.41} & 70.6 & 57.4 & 39.3 & \underline{70.6} & 54.5 & 42.4 & 25.3 & 33.3 & 67.16 \\
        SparseAdapter~\cite{he2022sparseadapter} & \underline{79.1} & 89.2 & 65.7 & 98.6 & \underline{89.3} & 68.5 & 52.5 & 77.58 & 79.5 & 94.7 & 79.4 & 74.4 & 81.99 & 70.2 & 56.9 & 37.8 & \underline{70.9} & 51.3 & 41.3 & 25.2 & 32.5 & 66.16 \\
        \rowcolor{pink} MoSA (Ours) & \textbf{79.7} & \textbf{91.5} & \underline{66.2} & \textbf{98.8} & \textbf{89.7} & \underline{79.0} & \textbf{53.4} & \textbf{79.86} & \textbf{83.4} & 95.6 & 82.0 & \underline{75.1} & \textbf{84.03} & \textbf{71.5} & 58.1 & 40.7 & 70.2 & 57.8 & 43.6 & \underline{26.5} & 34.0 & \underline{68.25} \\
        \bottomrule
        \end{tabular}
    }
\end{sidewaystable}

\textbf{Per-task results for MoSA on VTAB-1k.} \Cref{tab:vtab_detailed} shows the per-task results of MoSA on VTAB-1k. It can be seen that MoSA outperforms full fine-tuning on 13 tasks of VTAB, the highest among all PEFT methods. Additionally, MoSA also surpasses full fine-tuning (68.25\% \vs 65.57\%) and other baselines (AdaptFormer 67.16\%) in the average accuracy across 19 tasks.

\clearpage  

%
%
\bibliographystyle{splncs04}
\bibliography{main}
\end{document}